\documentclass[final]{cvpr}

\usepackage{times}
\usepackage{epsfig}
\usepackage{graphicx}
\usepackage{amsmath}
\usepackage{amssymb}
\usepackage{comment}
\newcommand{\bfsection}[1]{\vspace*{0.1cm}\noindent\textbf{#1.}}
\usepackage{multirow}
\usepackage[pagebackref=true,breaklinks=true,colorlinks,bookmarks=false]{hyperref}

\begin{document}

\title{Deep Lucas-Kanade Homography for Multimodal Image Alignment}

\author{Yiming Zhao \hspace{2cm} Xinming Huang
\hspace{2cm} Ziming Zhang\\
Worcester Polytechnic Institute\\

100 Institute Rd, Worcester, MA, USA\\

{\tt\small \{yzhao7, xhuang, zzhang15\}@wpi.edu}


}

\maketitle
\pagestyle{empty}
\thispagestyle{empty}

\begin{abstract}
  Estimating homography to align image pairs captured by different sensors or image pairs with large appearance changes is an important and general challenge for many computer vision applications. In contrast to others, we propose a generic solution to pixel-wise align multimodal image pairs by extending the traditional Lucas-Kanade algorithm with networks. The key contribution in our method is how we construct feature maps, named as deep Lucas-Kanade feature map (DLKFM). The learned DLKFM can spontaneously recognize invariant features under various appearance-changing conditions. It also has two nice properties for the Lucas-Kanade algorithm: (1) The template feature map keeps brightness consistency with the input feature map, thus the color difference is very small while they are well-aligned. (2) The Lucas-Kanade objective function built on DLKFM has a smooth landscape around ground truth homography parameters, so the iterative solution of the Lucas-Kanade can easily converge to the ground truth. With those properties, directly updating the Lucas-Kanade algorithm on our feature maps will precisely align image pairs with large appearance changes. We share the datasets, code, and demo video online \footnote{Codebase link: https://github.com/placeforyiming/CVPR21-Deep-Lucas-Kanade-Homography.  Dr. Xinming Huang is the project manager of this paper, please contact him for further information.}.
\end{abstract}

\section{Introduction}

\begin{figure}
  \centering
  \includegraphics[width=1.0\linewidth,height=50mm]{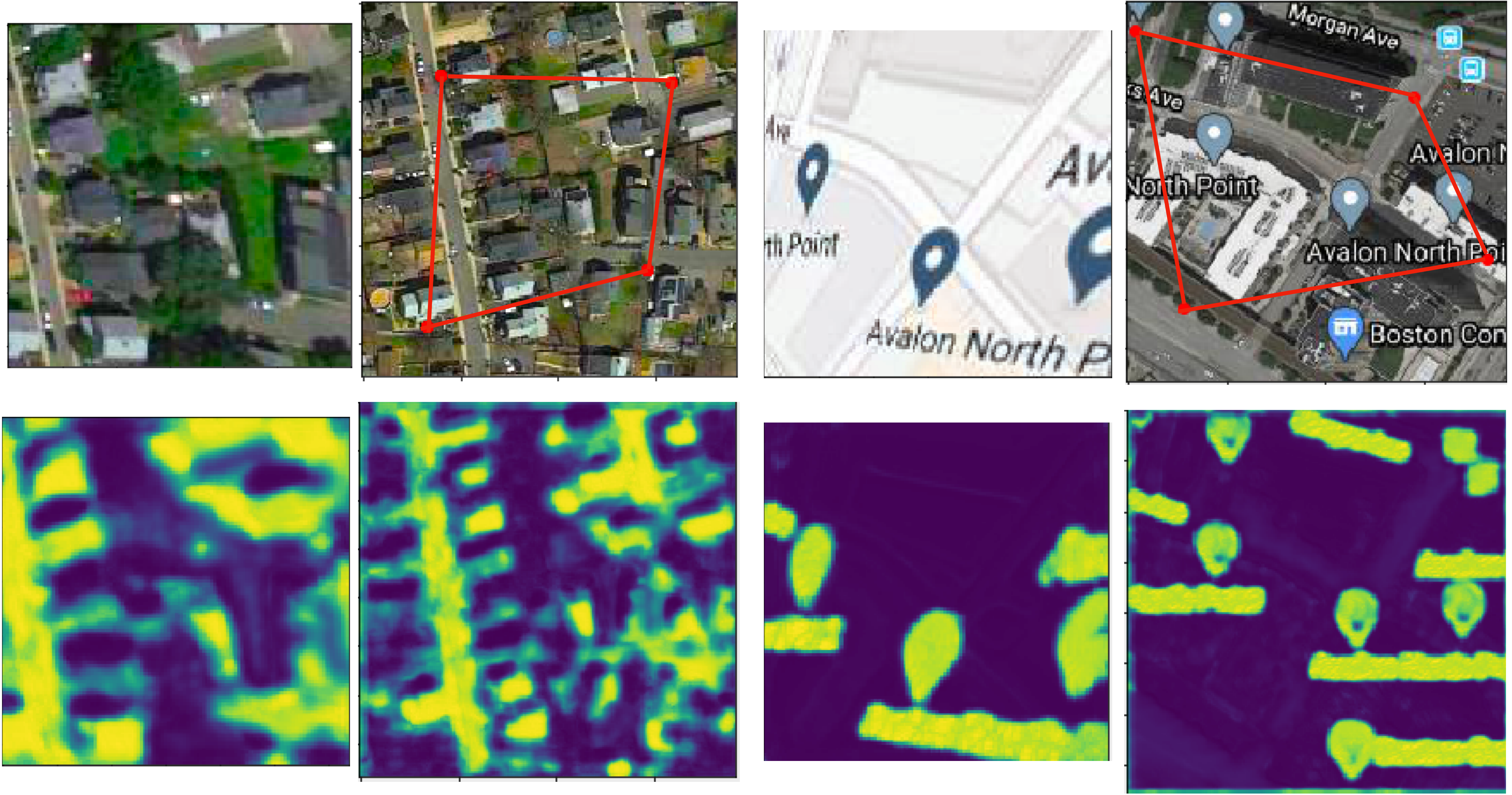}   
  \caption{{\bf (Left)} One image pair from our cross-season dataset captured in Google Earth. Plant and traffic are changing textures. Our DLKFM distinguishes invariant features which are roads and buildings. {\bf (Right)} One image pair from our Google Map and Satellite dataset. Template comes from the static map and the input is the corresponding satellite map. Our DLKFM distinguishes invariant features which are those map labels. The red polygon is the ground truth position of the template on the input image.}
\label{fig:1}
\vspace{-5mm}
\end{figure}

Pixel-wise alignment of multimodal image pair is an important problem for many computer vision applications in the field of medical imaging \cite{dolz2018hyperdense}, remote sensing \cite{ye2017robust}, and robotics \cite{cadena2016multi}. For example, in the task of GPS denied navigation \cite{goforth2019gps}, the drone needs to align the RGB image from the camera with the existing map from the satellite. Aligning those two images suffers from multimodal color representations as well as appearance changes. In Fig. \ref{fig:1}, we show a cross-season image pair with appearance changes and a Google Map and Satellite image pair with multimodal pattern representations. To solve this problem, we propose a generic pipeline by extending the traditional Lucas-Kanade (LK) algorithm with neural networks. The key component is how to extract the feature map, named as deep Lucas-Kanade feature map (DLKFM). DLKFM is able to spontaneously recognize invariant features in various multimodal cases, like the roads and houses in the cross-season example or map labels in the Google Map and Satellite example. DLKFM also helps the LK to converge successfully by shaping the landscape of the objective function.  

 Homography or perspective transform is a general 2D image transformation, which maps pixel from one image to the other, denoted as $\hat{x}=\hat{H}\hat{x^{\prime}},$ where $\hat{x}$ and $\hat{x^{\prime}}$ are homogeneous coordinates, $\hat{H}$ is an arbitrary $3\times 3$ matrix. However, in 3D computer vision, parallax arises due to the change of the viewpoint, thus creates ambiguity for aligning 2D images. In this paper, we consider the same flat world assumption as \cite{goforth2019gps}, that the depth disparity is negligible like some UAV and remote sensing applications \cite{nguyen2018unsupervised,zampieri2018multimodal}. Then, aligning the image pair is equivalent to estimating the homography matrix. 

To estimate the homography, a straightforward solution is finding more than four matched points. This idea can be broadly summarized as the feature-based method including traditional SIFT \cite{lowe2004distinctive}, SURF \cite{bay2006surf}, or recent deep learning methods D2 \cite{dusmanu2019d2}, LF-Net \cite{ono2018lf}, R2D2 \cite{revaud2019r2d2}, etc. To further solve the challenge on multimodal images, some special features are also designed \cite{ye2017robust, ye2019fast}. However, those feature-based methods are facing the same generalizability problem. Even for those learning-based features, ground truth labels provided by 3D reconstruction systems are necessary \cite{schonberger2016structure,schonberger2016pixelwise}. This prohibits the generic training of those feature detectors and descriptors on multimodal images.

Besides feature-based methods, direct method is the other way to calculate homography \cite{baker2004lucas}. As the most famous one, the Lucas-Kanade family \cite{lucas1981iterative} has a long history and still stimulates new ideas in different aspects of computer vision \cite{oron2014extended, wang2018deep}. In this paper, we build a generic multimodal image alignment pipeline by extending Inverse Compositional Lucas-Kanade (IC-LK). 

\bfsection{Inverse Compositional Lucas-Kanade Algorithm} Let $X_{I}$ and $X_{T}$ represent input image and template image, respectively, the Lucas-Kanade objective is stated as follows
\begin{align}\label{eq:1}
\min_{P} ||X_{T} -W (X_{I}, P)||^{2}_{2}
\end{align}
where $W(\cdot|P)$ is the warping function which warps the input image based on transformation parameter set $P$.

The Lucas-Kanade algorithm iteratively solves for the warp parameters $P_{k+1}=P_{k} + \Delta P$. At every iteration $k$, the warp increment $\Delta P$ is obtained by linearizing
\begin{align}\label{eq:2}
\min_{\Delta P} ||X_{T} -W (X_{I}, P_{k}+\Delta P) ||^{2}_{2}
\end{align}
using first order Taylor expansion
\begin{align}\label{eq:3}
\min_{\Delta P} ||X_{T} -W (X_{I}, P_{k})-\frac{\partial W(X_{I}, P_{k})}{\partial P}\Delta P ||^{2}_{2}
\end{align}
Here, $ \partial W(X_{I}, P_{k})/\partial P$ needs to be recomputed at every iteration as it depends on $W (X_{I}, P_{k})$. The inverse compositional(IC) \cite{baker2004lucas} avoids this recomputing by applying the warp increments $\Delta P$ to the template instead of the input
\begin{align}\label{eq:4}
\min_{\Delta P} ||W (X_{T},\Delta P) -W (X_{I}, P_{k}) ||^{2}_{2}
\end{align}
using the warp parameter update $P_{k+1}=P_{k} + (\Delta P)^{-1}$ as where $\Delta P$ is the inverse increment mapping. In the corresponding linearized equation
\begin{align}\label{eq:5}
\min_{\Delta P} ||X_{T}+\frac{\partial W(X_{T}, 0)}{\partial P}\Delta P -W (X_{I}, P_{k}) ||^{2}_{2}
\end{align}
$\partial W(X_{T}, 0)/\partial P$ does not depend on $P_{k}$ and can thus be pre-computed, resulting in a more efficient algorithm.

\bfsection{Limitation of IC-LK on Multimodal Images} IC-LK keeps accumulating $\Delta P$ toward ground truth, thus it usually has an accurate performance if it can work successfully. However, Eq. \ref{eq:1} assumes the color difference between the input image and template image achieves the minimum value once they are aligned. This brightness consistency assumption does not hold when aligning multimodal image pairs as the input and template usually have different color pattern representations. Furthermore, minimizing Eq. \ref{eq:1} is a non-linear optimization task as the input image depends non-linearly on the warp parameters $P$, so there is no guarantee the first-order approximation solution of Eq. \ref{eq:5} can have a $\Delta P$ moving along the right direction to the ground truth without getting into the local minimum.

\bfsection{Our Solution} To overcome the limitation of IC-LK on multimodal images, we hope feature maps processed after a Siamese network can satisfy those assumptions of IC-LK. To achieve this, we design a feature constructor inspired by classical feature extractors, such as Harris Corner \cite{derpanis2004harris}. This feature constructor constructs one channel feature map by using eigenvalues of the local covariance matrix on output tensor. Then, we design a special loss function with two terms. The first term helps feature maps to satisfy the brightness consistency assumption. The second term shapes the landscape of the Lucas-Kanade objective function, to help the first-order linearized solution $\Delta P$ can successfully keep moving toward the ground truth homography parameters. We show examples of learned deep Lucas-Kanade feature map (DLKFM) in Fig. \ref{fig:1}.

\bfsection{Contributions} In general, we extend the well-known traditional Lucas-Kanade method with neural networks. In regard to applications \cite{goforth2019gps,zampieri2018multimodal,nassar2018deep}, we propose a generic pipeline to align multimodal images. In regard to technical contributions, we develop two new specific technologies: 
\begin{itemize}
\item \emph{Feature constructor on multi-channel tensors.} In order to avoid updating homography on heavy tensors, the single-channel DLKFM is constructed on the high dimensional network output. We calculate the covariance matrix of vectors in each $3 \times 3$ patch. Then, we approximate the ratio between the largest eigenvalue and the trace as an indicator to form a single channel feature map. In the ablation study section, we show this operation improves the performance by extracting meaningful features on the single-channel feature map.  
\item \emph{Special loss function for the convergence of Lucas-Kanade.} We design a special loss function based on constructed feature maps. This loss function makes two feature maps satisfy brightness consistency. It further helps the convergence of the Inverse Compositional Lucas-Kanade algorithm by shaping the optimization objective with a supportive convex function. The landscape of the objective function on learned feature maps is visualized in the section of ablation study.      
\end{itemize}

\section{Related Work}

Image alignment is one of the most fundamental computer vision tasks with a long history \cite{baker2004lucas,szeliski2006image,goforth2019gps}. Direct methods \cite{lucas1981iterative}, feature-based methods \cite{liu2010sift} and their combinations \cite{antonakos2015feature} are traditional ways to align images. Those non-learning methods often assume input and template images have the same texture pattern for the same feature, which is not the case for multimodal images. To overcome this, some researchers try to design special feature detectors \cite{ye2017robust, ye2019fast}, which are usually hard to prove the generalizability. 

With the coming of the deep learning epoch, people are curious if the deep neural network can learn the corresponding features on two images \cite{long2014convnets,choy2016universal}. The positive answer attracts many recent research papers investigating how to utilize the strong ability of neural networks for image matching, like semantic matching or finding dense correspondences \cite{kim2018recurrent, kim2019semantic,truong2020glu}. Extending feature detectors and descriptors with the neural network is also a hot topic. Recent models, like D2 \cite{dusmanu2019d2}, LF-Net \cite{ono2018lf}, R2D2 \cite{revaud2019r2d2}, RF-Net \cite{shen2019rf}, Elf \cite{benbihi2019elf} are all showing strong ability to handle struggling cases for traditional feature detectors. However, many of those methods rely on ground truth feature points to train the network. Those ground truth feature points are traditional features extracted by 3D reconstruction systems, like Colmap \cite{schonberger2016structure,schonberger2016pixelwise}. This limits the training of those methods on multimodal datasets.

A more straightforward idea is making use of the strong approximation ability of networks to directly regress homography parameters \cite{detone2016deep, erlik2017homography}. Those deep homography papers usually focus on specific challenges, like how to design the unsupervised loss \cite{nguyen2018unsupervised}, or how to handle dynamic scenes \cite{le2020deep,zhang2020content}. Although using networks to do regression is simple, it always can work under various conditions. However, the model accuracy is usually not guaranteed.   

In contrast to above methods, we extend the traditional Inverse Compositional Lucas-Kanade method on multimodal image pairs. Two recent papers explore Lucas-Kanade with neural networks \cite{chang2017clkn,lv2019taking}. Authors of one paper replace some parts of the inverse compositional solution with trainable data-driven network modules \cite{lv2019taking}. However, their method mainly focuses on solving the 3D rigid motion estimation problem. In CLKN \cite{chang2017clkn}, authors design a loss function to guarantee the first updating always bring relative large progress, the further iterative updating is conducted on feature maps. Since there is no special treatment for feature maps, they imply the same color pattern will become the same feature via same network parameters. This assumption does not work for multimodal image pairs as two images have different color textures for the same place.     

\section{Method}

\subsection{Problem Settings}
To overcome the limitation of Lucas-Kanade on multimodal image pairs, we extend the Inverse Compositional Lucas-Kanade on feature maps. A Siamese network, which has one template branch defined as $F(\cdot|w_{T})$, and one input branch defined as $G(\cdot|w_{I})$, is used to extract feature maps. Then, the objective function in Eq. \ref{eq:1} can be generalized as: 

\begin{align}\label{eqn:braodcast_equation}
\min_{P} ||F(X_{T},\bar{w_{T}}) -W (G(X_{I},\bar{w_{I}}), P) ||^{2}_{2} .
\end{align}
Here, we have two kinds of set of parameters, the Siamese network parameter set $\{w_{T},w_{I}\}$, and homography parameter set $P$. During training, we use ground truth $\bar{P}$ to train $\{w_{T},w_{I}\}$, then the fixed network weights $\{\bar{w_{T}},\bar{w_{I}}\}$ are used to update $P$ during inference to minimize Eq. \ref{eqn:braodcast_equation}. The iteration inverse compositional solution of Eq. \ref{eqn:braodcast_equation} is 
\begin{align}\label{eqn:solution}
\Delta P=(J^{T}J)^{-1}J^{T}r,
\end{align}
where
\begin{align}\label{eqn:J}
J= \nabla F(X_{T},\bar{w_{T}})\cdot \frac{\partial W}{\partial P}(W(F(X_{T},\bar{w_{T}}),0), 
\end{align}
\begin{align}\label{eqn:r}
r = \begin{bmatrix}
W (G(x_{I},\bar{w_{I}}), P)_{1} - F(X_{T},\bar{w_{T}})_{1} \\
...\\
...\\
W (G(X_{I},\bar{w_{I}}), P)_{N} - F(X_{T},\bar{w_{T}})_{N} 
\end{bmatrix}
\end{align}
The derive details from Eq. \ref{eqn:braodcast_equation} to Eq. \ref{eqn:solution} can be found from equation (2) to (8) here \cite{chang2017clkn}. Since the updating of Eq. \ref{eqn:solution} is conducted on trained feature maps during inference, we use $\{\bar{w_{T}},\bar{w_{I}}\}$ to represent fixed network weights during inference. In Eq. \ref{eqn:r}, $N$ is the total number of elements on the feature maps. 

\subsection{Construct Single Channel Feature Map}

The modern convolution layers usually output a high dimension tensor with multiple filters, like 64, 128, or more. This involves a large computational burden to update $\Delta P$ in Eq. \ref{eqn:solution} and Eq. \ref{eqn:r}. Some recent papers, which also try to involve neural networks into the pipeline of Lucas-Kanade, usually take straightforward choices like reducing the number of filters in the last layer to a small value \cite{chang2017clkn}, or applying channel-wise summation on the output tensor \cite{lv2019taking}. In contrast to them, we construct a single channel feature map based on eigenvalues of the local covariance matrix.

Considering a large tensor $B$, we apply a $3 \times 3$ sliding window to extract 9 vectors locally. Then, we calculate the covariance matrix of those 9 vectors. We use $B_{i,j}$ to denote the covariance matrix calculated by the local patch centered at $(i,j)$ of tensor $B$. Intuitively, if the information captured by this local patch is almost the same, those 9 vectors should be similar. Thus, the covariance matrix will not have a dominant eigenvalue. So we construct a one channel feature map $C$ by using the ratio of the largest eigenvalue over the summation of all eigenvalues:
\begin{align}\label{eqn:trace}
C_{i,j}=\frac{\lambda_{largest}}{Trace(B_{i,j})},
\end{align}
where $\lambda_{largest}$ is the largest eigenvalue of $B_{i,j}$.
In traditional image processing, we as humans think corners or edges are important features for an image, like the Harris corner detector \cite{derpanis2004harris} or Canny edge detector \cite{ding2001canny}. In many different subjects which need matrix analysis, like statistics and signal processing, the ratio of largest eigenvalue over trace plays an important role \cite{nadler2011distribution}. So we think this ratio is an important feature for high dimensional tensor, just like the edge is an important feature for the image.  

However, we found calculating and doing backpropagation on the largest eigenvalue for each patch is super slow with modern deep learning frameworks. Then, we design a more practical indicator by following Perron–Frobenius theorem \cite{perron1907theorie,frobenius1912matrizen}. According to Perron–Frobenius theorem, the largest eigenvalue of the covariance matrix has an upper bound which is the maximum of summation of each row, and a lower bound which is the minimum of summation of each row. So the new feature map of a high dimension tensor is constructed by:  
\begin{align}\label{eqn:one_channel}
C_{i,j}=\frac{max_{m}(\sum_{n}B_{i,j}^{m,n})+min_{m}(\sum_{n}B_{i,j}^{m,n})}{2\times Trace(B_{i,j})}.
\end{align}
where $m, n$ are row and column numbers of the covariance matrix. The calculation of this feature map is just like doing regular convolution, which keeps our method practical for training and testing. We discuss more details in the section of the ablation study.

\subsection{Build Loss Function} 
\begin{figure*}
  \centering
  \includegraphics[width=0.95\linewidth, height=55mm]{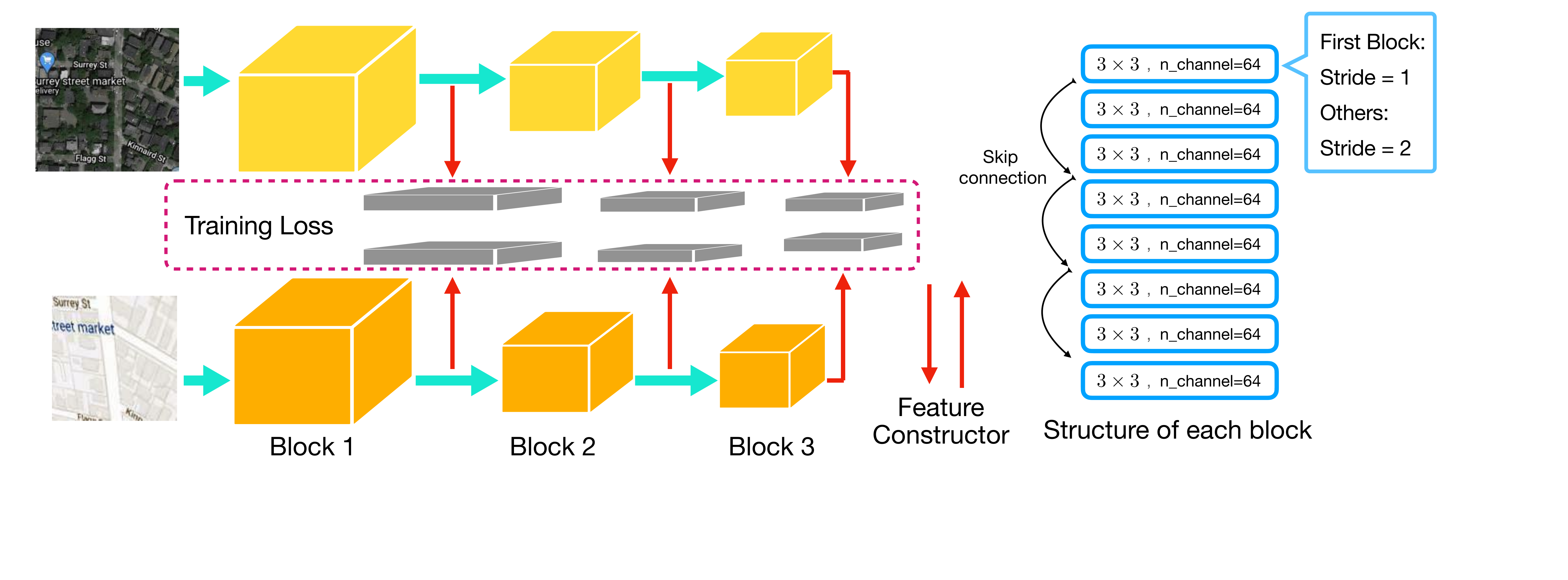}   
  \caption{The Siamese network has three identical blocks except for the stride in the first layer. The first layers in the second and third block have stride 2 to downsample the input tensor. We construct one channel feature maps on the output tensors of each block. The red arrow represents the feature constructor developed in this paper. Each block contains a few layers with residual skip connection. The loss is built on constructed feature maps during training.}
\label{fig:2}
\vspace{-1mm}
\end{figure*}

\bfsection{Brightness Consistency Loss Term} Once we have two feature maps from the template branch and the input branch in the Siamese network, we need to design a loss function to help those two feature maps learn meaningful features for Eq. \ref{eqn:solution}. The original Lucas-Kanade objective Eq. \ref{eq:1} assumes that the color difference should be the minimum once two images are well aligned. We apply this assumption on feature maps to get the first term of our loss function named Brightness Consistency Loss $l_{bc}=E(w_{T},w_{I}|\bar{P})$, where
\begin{align}\label{eqn:loss_1}
E(w_{T},w_{I}|\bar{P})=||F(X_{T},w_{T}) -W (G(X_{I},w_{I}), \bar{P})||^{2}_{2} 
\end{align}
Compared with Eq. \ref{eqn:braodcast_equation}, $l_{bc}$ has trainable $\{w_{T},w_{I}\}$ with fixed ground truth $\bar{P}$ on training set. The warping function is implemented by using STN \cite{jaderberg2015spatial} for backpropagation.  

\bfsection{Convergence Loss Term} Original objective Eq. \ref{eq:1} is a highly non-linear function, whereas the iteration solution Eq. \ref{eqn:solution} relies on gradient descent approximation. To help it converge better in the right direction, we hope the objective Eq. \ref{eqn:braodcast_equation} on feature maps can achieve local minimum at ground truth $\bar{P}$ and have a smooth surface around $\bar{P}$ without the other minimum. Control the convexity of a function usually needs to explicitly calculate Jacobian or Hessian, but Eq. \ref{eqn:braodcast_equation} is a multivariable function of $P$ with a Siamese network as the backbone. Instead of directly calculating Jacobian or Hessian, we consider a supportive convex function $g(Z)$ with $J(g(Z))|_{Z=\bar{P}}=0$ to shape the local region of objective function around $\bar{P}$. The $Z$ is a vector with eight parameters and $g(Z)$ has the minimum value at $Z=\bar{P}$. We push the optimization objective function Eq. \ref{eqn:braodcast_equation} satisfy two conditions in a small region $\Theta$ around $\bar{P}$ by training:
$$ Condition 1: {\forall} (\bar{P}+\Delta P) \in \Theta, $$
$$E(\bar{P}+\Delta P|\bar{w_{T}},\bar{w_{I}})-E(\bar{P}|\bar{w_{T}},\bar{w_{I}})\geq g(\bar{P}+\Delta P)-g(\bar{P}). $$  
This condition guarantees that Eq. \ref{eqn:braodcast_equation} achieves local minimum at $\bar{P}$ with a relatively larger local change than $g(Z)$.
$$ Condition 2: {\forall} P \in \Theta , \nabla_{v} E(P|\bar{w_{T}},\bar{w_{I}})\geq \nabla_{v} g(P),$$ 
where $\nabla_{v}$ is the directional derivative. Since \emph{Condition 1} is about the single point at $\bar{P}$, \emph{Condition 2} guarantees that Eq. \ref{eqn:braodcast_equation} has a steeper directional derivative than $g(Z)$ in a local region around $\bar{P}$. Compared with partial derivative, directional derivative is easier to be implemented for function without explicit expression. 

In our implementation, we use the simplest $g(Z)=\sum_{i=1}^{8}(Z_{i}-\bar{P}_{i})^2$ as the supportive function. Then, the loss term for \emph{Condition 1} can be defined as:
\begin{align}
\begin{split}
l_{con1} & =-\frac{1}{M}\sum_{m=1}^{M} minimum(0, E(\bar{P}+\Delta P_{m}|w_{T},w_{I}) \\
& - E(\bar{P}|w_{T},w_{I})- \sum_{i=1}^{8}(\Delta P_{i}^{m})^2),
\end{split}
\label{eqn:loss_2}
\end{align}
where $\Delta P_{i}^{m}$ is a sampled random noise of the $i$-th variable in vector $P$, we sample $M$ times for each training batch.

For \emph{Condition 2}, we define ${\forall} (\bar{P}+\Delta P) \in \Theta$,  $P=\bar{P}+\lambda \Delta P$, where $\lambda \in (0,1)$. Then, we have:
\begin{align}\label{eqn:loss_3}
\begin{split}
&\nabla_{\Delta P} E(P|\bar{w_{T}},\bar{w_{I}}) =  \lim_{(1-\lambda) \to 0} \\
& \frac{E(\bar{P}+\Delta P|\bar{w_{T}},\bar{w_{I}})-E(\bar{P}+\lambda \Delta P|\bar{w_{T}},\bar{w_{I}})}{(1-\lambda)||\Delta P||},
\end{split}
\end{align}
\begin{align}\label{eqn:loss_4}
\nabla_{\Delta P} g(P)& =  \lim_{(1-\lambda) \to 0} \frac{\Delta P^2- (\lambda \Delta P)^2}{(1-\lambda)||\Delta P||},
\end{align}
Combining Eq. \ref{eqn:loss_3} and Eq. \ref{eqn:loss_4}, we have the mathematical expression of \emph{Condition 2} as
\begin{align}\label{eqn:loss_5}
\begin{split}
&{\forall} (\bar{P}+\Delta P) \in \Theta, E(\bar{P}+\Delta P|\bar{w_{T}},\bar{w_{I}}) \\
&-E(\bar{P}+\lambda \Delta P|\bar{w_{T}},\bar{w_{I}}) \geq (1-\lambda^2)\sum_{i=1}^{8}(\Delta P_{i}^{m})^2 .
\end{split}
\end{align}
The loss term for \emph{Condition 2} can be defined as :
\begin{align}\label{eqn:loss_6}
\begin{split}
&l_{con2}=-\frac{1}{M}\sum_{m=1}^{M} minimum(0, E(\bar{P}+\Delta P_{m}|w_{T},w_{I})\\
&-E(\bar{P}+\lambda \Delta P_{m}|w_{T},w_{I})- (1-\lambda^2)\sum_{i=1}^{8}(\Delta P_{i}^{m})^2).
\end{split}
\end{align}

Since we have $\lambda \in (0,1)$ and $(1-\lambda) \to 0$, we choose $\lambda=0.8$ for the implementation.

Finally, the Convergence Loss is defined as $l_{converge}=l_{con1}+l_{con2}$. The total loss function is the summation of the Brightness Consistency Loss and the Convergence Loss
\begin{align}\label{eqn:final_loss}
l_{total}= l_{bc}+\gamma l_{converge}
\end{align}
where $\gamma$ is a hyperparameter to control the balance. We choose $\gamma=0.1$ in this paper.

\subsection{Network Structure}
We show the network structure in Fig. \ref{fig:2}. The Siamese network has three identical blocks except for the first layer. In each block, there are 8 convolution layers with 64 filters connected with the residual. The first layers in the second and third blocks have stride 2 to downsample tensors. All the other layers have $3\times 3$ kernels with stride 1.
 
During training, the input image and template image go through two branches. Then, we extract output tensors after each block, and apply Eq. \ref{eqn:one_channel} to construct feature maps for both input and template. The loss function Eq. \ref{eqn:final_loss} is defined based on constructed feature maps in three different scales. 

\begin{figure*}
  \centering
  \includegraphics[width=0.95\linewidth, height=40mm]{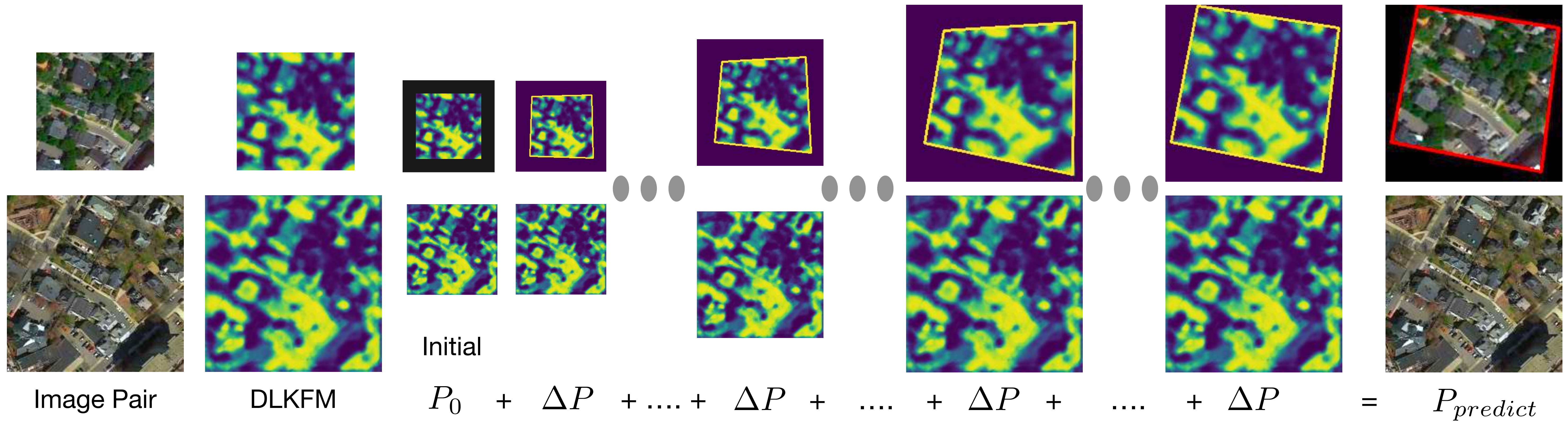}   
  \caption{For a given initial guess, the Eq. \ref{eqn:solution} is used to calculate $\Delta P$ on input and template DLKFM for updating $P$. The updating starts from the smallest scale with a certain stop criterion on each scale. More examples of iterative updating are shown in the demo video.}
\label{fig:3}
  \vspace{-2mm}
\end{figure*}

During inference, we have feature maps in three different scales. The Eq. \ref{eqn:solution} is updated bottom-up from coarse to fine. We give an example of the updating during inference in Fig. \ref{fig:3}. Note that the iterative solution needs an initial guess to start with and a stopping criterion to end, we discuss those two in the section of experiments.

\section{Experiments}

\subsection{Datasets}

To test the performance of our method, we prepare three datasets. We utilize Adam \cite{kingma2014adam} as our optimizer with a learning rate 0.0001 to train 10 epochs for all three datasets. As discussed in the section of method, we set $\gamma = 0.1$, $\lambda = 0.8$ and $M = 4$. All the experiments are conducted on a single RTX 2080 Ti.

\begin{figure}
  \centering
  \includegraphics[width=0.95\linewidth, height=60mm]{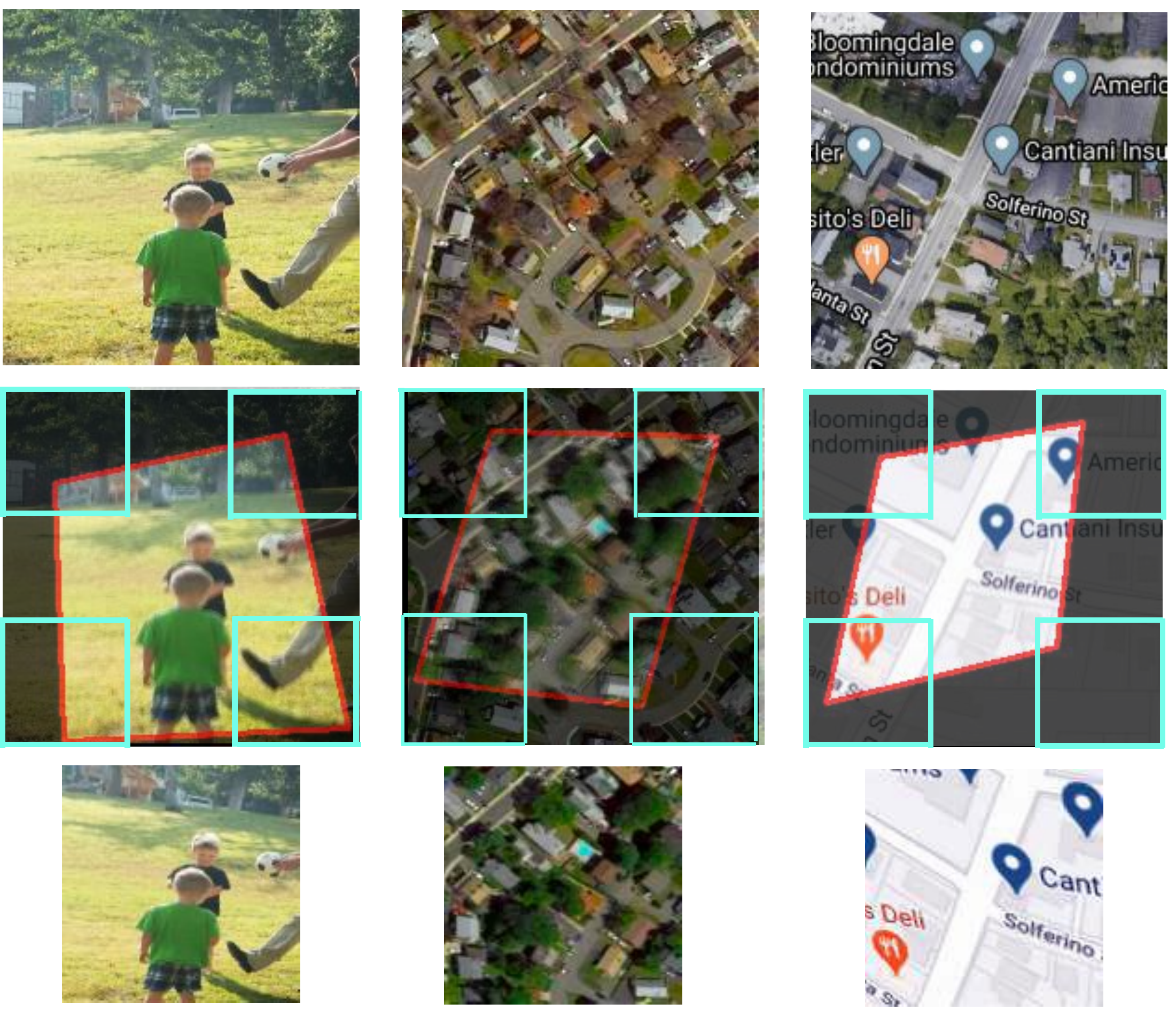}   
  \caption{From left to right: MSCOCO, Google Earth cross season dataset, Google static map and satellite dataset. We show how to prepare the input and template images for those three datasets. }
\label{fig:4}
\vspace{-2.em}
\end{figure}

\textbf{MSCOCO} \cite{lin2014microsoft} is a widely used large scale dataset. Recent deep homography papers also choose this dataset as a test bench. We exactly follow the settings of those papers \cite{detone2016deep,chang2017clkn,le2020deep}. The whole training set of MSCOCO is used for our training, and we sample 6K images from the validation set as our test. Each sample is resized to $196\times 196$ as an input image. Then, we randomly choose four points in four $64 \times 64$ boxes at the corner and warp the chosen region to $128 \times 128$ as the template. An example is showed in Fig. \ref{fig:4}. The use of this regular dataset is serving for two purposes. First, since we implement several baseline methods, including DHN \cite{detone2016deep}, MHN \cite{le2020deep}, SIFT with Ransac, etc, observing the performance drop from the same model on multimodal datasets will illustrate the challenge of estimating homography on multimodal images. Second, we want to show our method can also work on the regular image alignment task.

\textbf{Google Earth} provides high-resolution satellite images captured on different dates. So, we can choose images for the same place captured in different seasons. Here we are using satellite images saved on 04/2018 and 06/2019 about the Great Boston area. One sample is provided in the second row of Fig. \ref{fig:4} to show that input and template images have appearance changes like different traffic and plant conditions. We randomly sample around 8K images as training, and around 1K as testing. Since randomly sampling will choose some completely textureless regions, like the surface of the river, we manually remove some images to finally get 850 test samples. 

\textbf{Google Maps and Satellite} are multimodal images provided by Google Static Map API. Two corresponding images belong to static google maps and satellite maps respectively. One example is shown in the third row of Fig. \ref{fig:4}, we can see they have completely different color pattern representations of the same place. We randomly sample around 8K images in the same place as training. For testing, we follow the same way as cross-season Google Earth dataset, to prepare 888 sample pairs finally.   

\subsection{Evaluation Metric}

The same as recent papers \cite{chang2017clkn,le2020deep}, we are using average corner error in pixel as the evaluation metric:
\begin{align}\label{eqn:metric}
e_{c}(P,\bar{P})=\frac{1}{4}\sum_{j=1}^{4}|| W(e_{j},P)-W(e_{j},\bar{P})||_2,
\end{align}
where $e_1,...,e_4$ are four corners in template image.
The corner error measures the $L_{2}$ distance of the warped corners and then takes the average over the four corners.

\begin{figure*}
  \centering
  \includegraphics[width=0.99\linewidth, height=47mm]{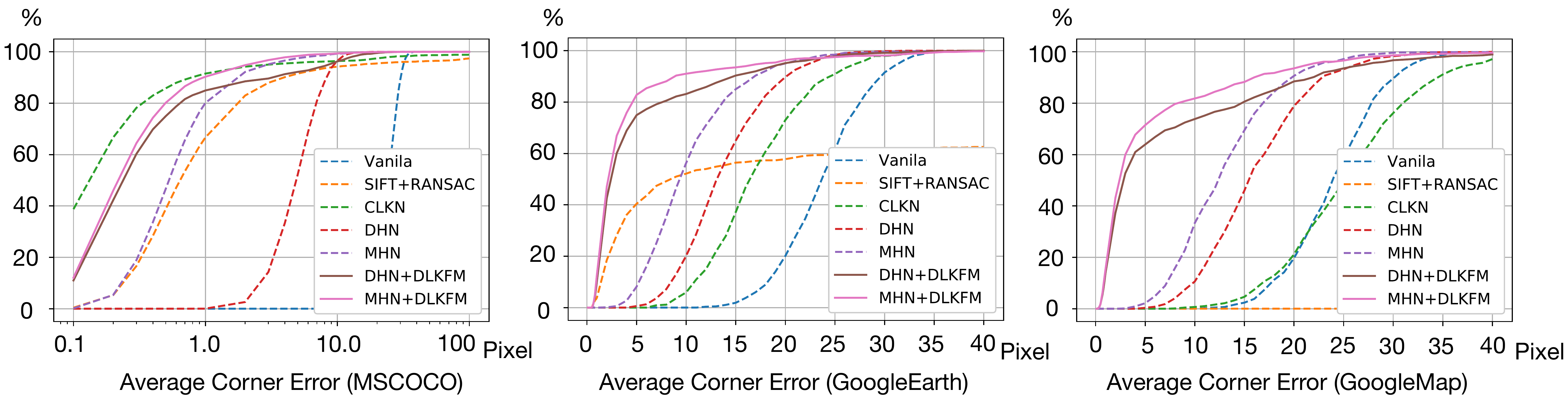}   
  \caption{The x-axis is the mean average pixel error, the y-axis is the cumulative percentage of test images which have lower average pixel error than x. Since almost all methods work well on MSCOCO, the x-axis is in a log scale for this dataset. We use two regression methods DHN \cite{detone2016deep} and MHN \cite{le2020deep} to provide the initial $P_{0}$ as they can work on all three datasets. It is clear to see that updating Eq. \ref{eqn:solution} on our deep Lucas-Kanade feature map (DLKFM) significantly improves the alignment accuracy. }
\label{fig:5}
  \vspace{-1.em}
\end{figure*}

\subsection{Initial Homography}

Updating Eq. \ref{eqn:solution} on DLKFM needs an initial homography $P_{0}$. The initial used in Fig. \ref{fig:3} is a pre-assigned homography matrix that we simply set (31,31), (31,159), (159,159), (159,31) as four corners of a $128 \times 128$ template image on $192 \times 192$ input. Note, the closer the initial homography to the ground truth, the larger chance Lucas-Kanade will converge to it. In some applications, like tracking \cite{oron2014extended} or navigation \cite{chao2014survey}, the previous frame can serve as the initial candidate since the relative pose is small. Some traditional methods, like fast affine template matching \cite{korman2013fast}, can also work to provide an initial guess with the brightness consistency property of our feature maps. 

In this paper, we choose two recent network-based deep homography models, DHN \cite{detone2016deep} and MHN \cite{le2020deep}, to respectively provide the initial homography. This helps our solution achieve state-of-the-art performance on recent benchmark settings \cite{chang2017clkn,le2020deep}. It is also able to show the strong ability of our method to work as a post-process refiner that keeps improving the alignment accuracy based on existing or future solutions. 

\subsection{Stopping Criterion} 
The iterative solution Eq. \ref{eqn:solution} needs a stopping criterion during the inference. For each $\Delta P$ on $P_{k}$, the average four corner error can still be used to evaluate relative improvement of $\Delta P$ as $e_{c}(P_{k}+\Delta P,P_{k})$. In this paper, we set the stopping criterion as $e_{c}(P_{k}+\Delta P,P_{k})<1$ pixel for smallest scale, $e_{c}(P_{k}+\Delta P,P_{k})<0.1$ pixel for middle scale, and $e_{c}(P_{k}+\Delta P,P_{k})<0.01$ pixel for largest scale. To further control the inference time, the maximum iteration number on each scale is 30.

\subsection{Model Performance}
We show the performance of our method and all baseline methods in Fig. \ref{fig:5}. In general, our solution outperforms the others on multimodal datasets with a large margin. We give further discussions about each method below.  

\bfsection{Vanilla} Based on how we generate templates, the vanilla means centers of four corner boxes are used as the prediction. This is a baseline for guessing without knowledge.

\bfsection{SIFT+RANSAC} This is a typical traditional method to calculate homography. We use OpenCV implementation of SIFT, BF Matcher, and RANSAC. From Fig. \ref{fig:5}, we can see it achieves pretty accurate performance on the MSCOCO dataset which has the exact same template and input images. However, on cross-season Google Earth dataset, it performs better than vanilla for only 60\% of images, and it completely fails to calculate homography on Google Maps. 

\begin{figure}[h]
  \centering
  \includegraphics[width=0.99\linewidth, height=35mm]{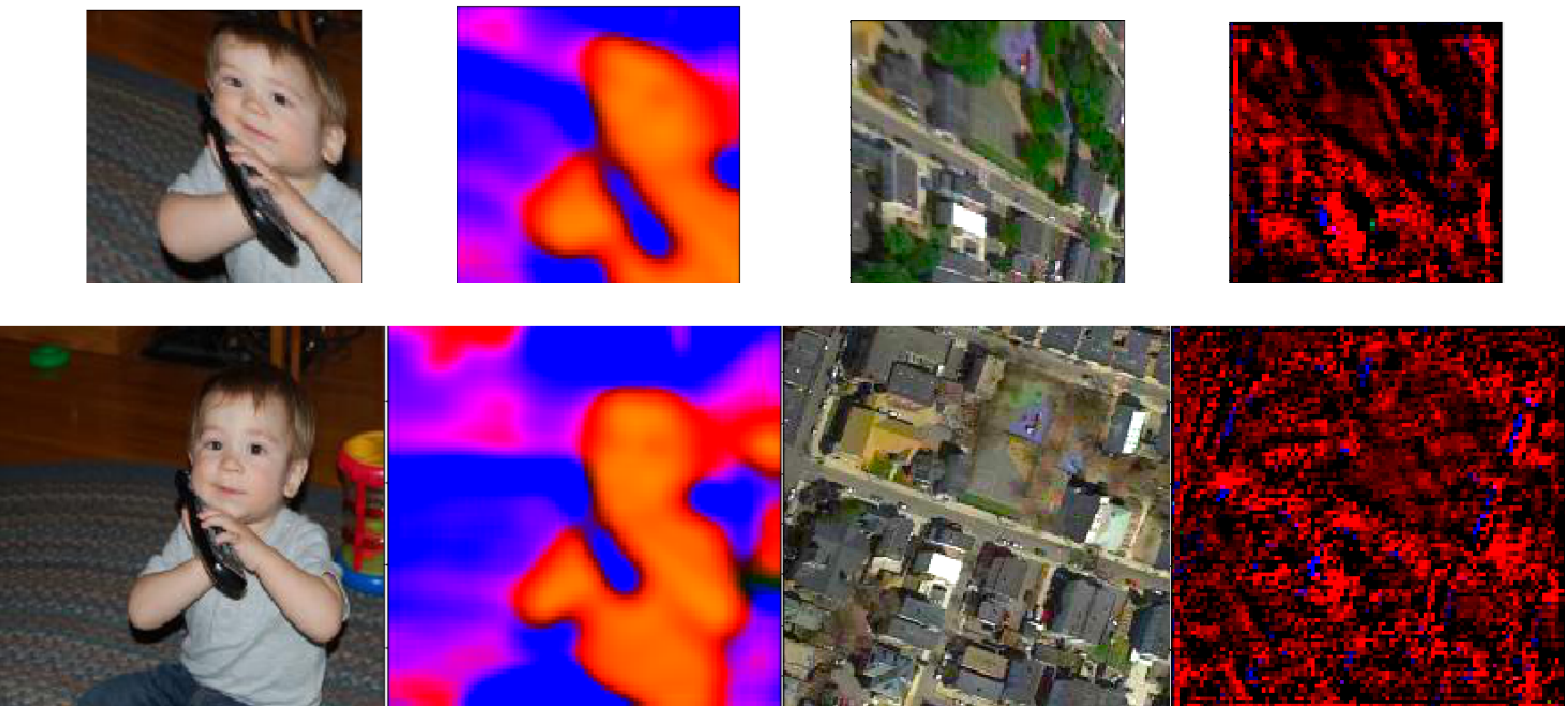}   
  \caption{This figure illustrates why CLKN \cite{chang2017clkn} can not work on multimodality case. {\bf (Left)}: the images and learned feature maps of CLKN \cite{chang2017clkn} on MSCOCO, we can see that feature maps extract meaningful semantic information. {\bf (Righ)}: the images and learned feature maps on cross season Google Earth, we can see the learned feature maps are no longer meaningful.}
\label{fig:7}
  \vspace{-1.0em}
\end{figure}

\bfsection{CLKN \cite{chang2017clkn}} This method extends inverse compositional Lucas-Kanade on feature maps. Based on vanilla as initial, it relies on training to help the first updating get relatively large progress. However, since CLKN \cite{chang2017clkn} does not set any constrain on feature maps, it fails to align multimodal images. We show the feature maps learned by CLKN on the MSCOCO dataset and the Google Earth cross-season dataset respectively in Fig. \ref{fig:7}. It is clear to see learned feature maps on multimodal images are not meaningful.

\bfsection{DHN \cite{detone2016deep} and MHN \cite{le2020deep}} Those two methods are using the network as a regression function to directly output homography parameters. A recent technical report gives a comprehensive comparison between those two methods and many traditional methods under various conditions  \cite{niblick2020homography}. MHN \cite{le2020deep} is a more advanced version as it combines three DHN \cite{detone2016deep} in a multi-scale cascaded way. Though the idea is straightforward, we find both two methods always can give meaningful results for different cases. However, compared with the regular MSCOCO dataset, we observe the performance drop of those two methods on multimodal datasets showed in Table \ref{tab:1}. For example, MHN \cite{le2020deep} performs much better than DHN \cite{detone2016deep}, and aligns 95\% of total test images on MSCOCO within 3-pixel error, but it fails to align almost all test multimodal image pairs within 3 average pixels.  

\begin{table}[ht]
\begin{center}
\begin{tabular}{c|c|c|c}
 \hline
  Dataset & Method & \textless 3 pixel& \textless 10 pixel \\
    \hline
    \multirow{4}{*}{MSCOCO}&DHN\cite{detone2016deep} &14\%&97\%\\
    \multirow{4}{*}&DHN+DLKFM&90\%&97\%\\\cline{2-4}
    \multirow{4}{*}&MHN\cite{le2020deep}&95\%&99\%\\
    \multirow{4}{*}&MHN+DLKFM&97\%&99\%\\
    \hline
    \multirow{4}{*}{GoogleEarth}&DHN\cite{detone2016deep} &0\%&20\%\\
    \multirow{4}{*}&DHN+DLKFM&60\%&83\%\\\cline{2-4}
    \multirow{4}{*}&MHN\cite{le2020deep}&1\%&56\%\\
    \multirow{4}{*}&MHN+DLKFM&66\%&91\%\\
    \hline
    \multirow{4}{*}{GoogleMap}&DHN\cite{detone2016deep} &0\%&10\%\\
    \multirow{4}{*}&DHN+DLKFM&52\%&74\%\\\cline{2-4}
    \multirow{4}{*}&MHN\cite{le2020deep}&0\%&33\%\\
    \multirow{4}{*}&MHN+DLKFM&60\%&81\%\\
    \hline
\end{tabular}
\end{center}
  \vspace{-0.5em}
\caption{The percentage of test images with a smaller average corner error than 3 and 10 pixels. Our method largely improves the performance of existing methods by using them as initials.}
\label{tab:1}
  \vspace{-0.5em}
\end{table}

\bfsection{DHN + DLKFM and MHN + DLKFM}  As our major contribution, DLKFM guarantees the local minimum and local smoothing of the Lucas-Kanade objective function. The iterative updating solution of Eq. \ref{eqn:solution} needs the initial homography to start with. As discussed before, we choose results from regression models DHN \cite{detone2016deep} and MHN \cite{le2020deep} as the initial $P_{0}$ respectively. This makes our pipeline generic, which is not limited to the experimental settings in this paper. According to Fig. \ref{fig:5} and one recent technical report \cite{niblick2020homography}, those regression models are always able to give relatively good performance under different conditions. In Table. \ref{tab:1}, we show the improvement by updating Eq. \ref{eqn:solution} on DLKFM. On multimodal datasets, our method largely improves the accuracy by helping most of the image pairs get smaller average corner errors than 3 pixels. On MSCOCO, even though MHN \cite{le2020deep} is already very accurate, we can still improve it. 

\section{Ablation Study}

We proposes two specific technologies in this paper. The feature constructor constructs one channel feature map based on a high dimension tensor. The special designed loss function guarantees the success of Inverse Compositional Lucas-Kanade on learned feature maps. We verify those two technical contributions here.       

\subsection{Effectiveness of Feature Constructor}

\begin{figure}[t]
  \centering
  \includegraphics[width=0.95\linewidth, height=27mm]{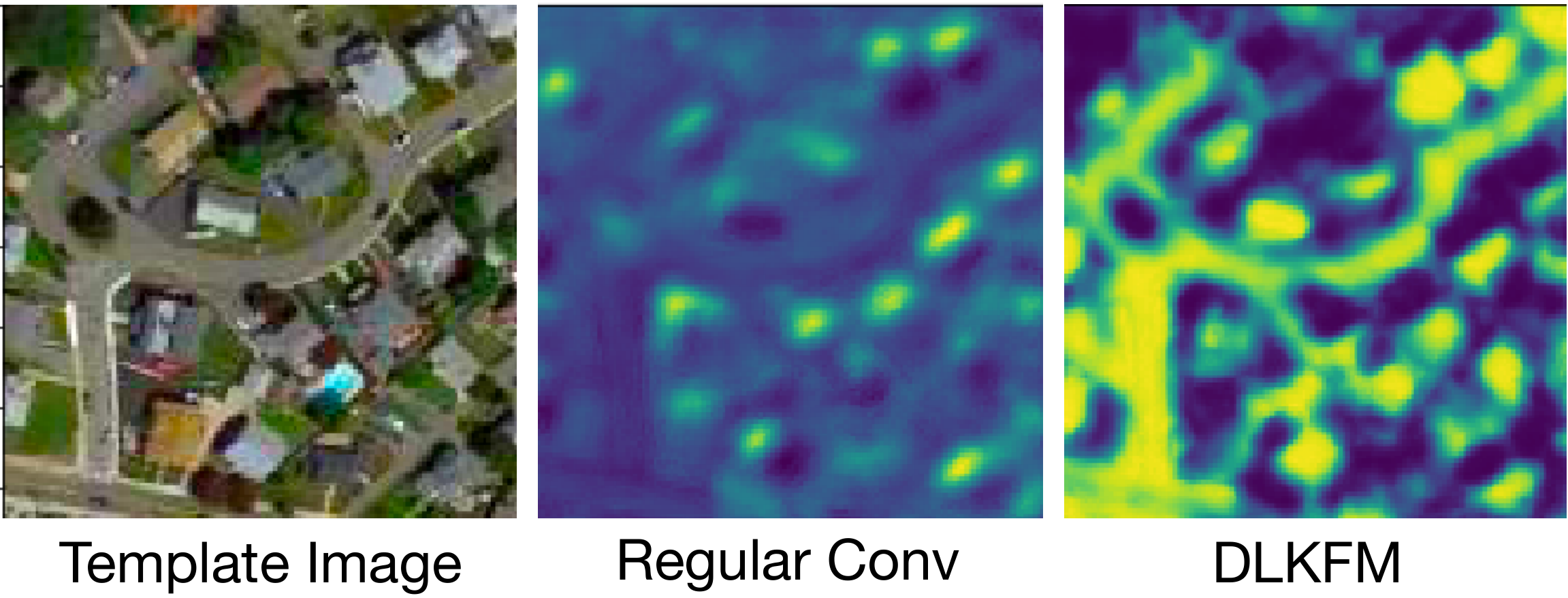}   
  \caption{The demonstration of the meaningful feature map extracted by our feature constructor, DLKFM. }
\label{fig:6}
  \vspace{-1.5em}
\end{figure}

\begin{table}[ht]
\begin{center}
\begin{tabular}{c|c|c|c}
 \hline
  Initial Method & Feature type  &\textless 3 pixel& \textless 10 pixel \\
\hline
    \multirow{2}{*}{DHN \cite{detone2016deep}}& Regular Conv &47\%&72\%\\
    \multirow{2}{*}&DLKFM&60\%&83\%\\
    \hline
    \multirow{2}{*}{MHN \cite{le2020deep}}&Regular Conv&55\%&83\%\\
    \multirow{2}{*}&DLKFM&66\%&91\%\\
    \hline
\end{tabular}
\vspace{-0.5em}
\end{center}
\caption{Performance comparison on cross season Google Earth.}
\label{tab:3}
  \vspace{-0.8em}
\end{table}

Since Eq. \ref{eqn:J} and Eq. \ref{eqn:r} need to be calculated with all values from the output tensor, we reduce the number of channels to one by using Eq. \ref{eqn:one_channel}. The other common way is using a regular convolutional layer with fewer filters \cite{chang2017clkn,lv2019taking}. We conduct an extra experiment on cross-season Google Earth dataset by replacing our feature constructor with a regular one filter convolutional layer. We list the comparison performance in Table \ref{tab:3}. To further demonstrate the effectiveness of our feature constructor, we visualize the learned feature map in Fig. \ref{fig:6}. It is clear that our feature constructor helps the feature map extract meaningful features, which are invariant roads and houses. In recent papers \cite{zampieri2018multimodal,nassar2018deep}, those features are learned with semantic labels, like the building label provided by OpenStreetMap \cite{haklay2008openstreetmap}, whereas our method spontaneously recognizes those patterns.

\subsection{Effectiveness of Designed Loss Function}

\begin{figure}[t]
  \centering
  \includegraphics[width=0.99\linewidth, height=40mm]{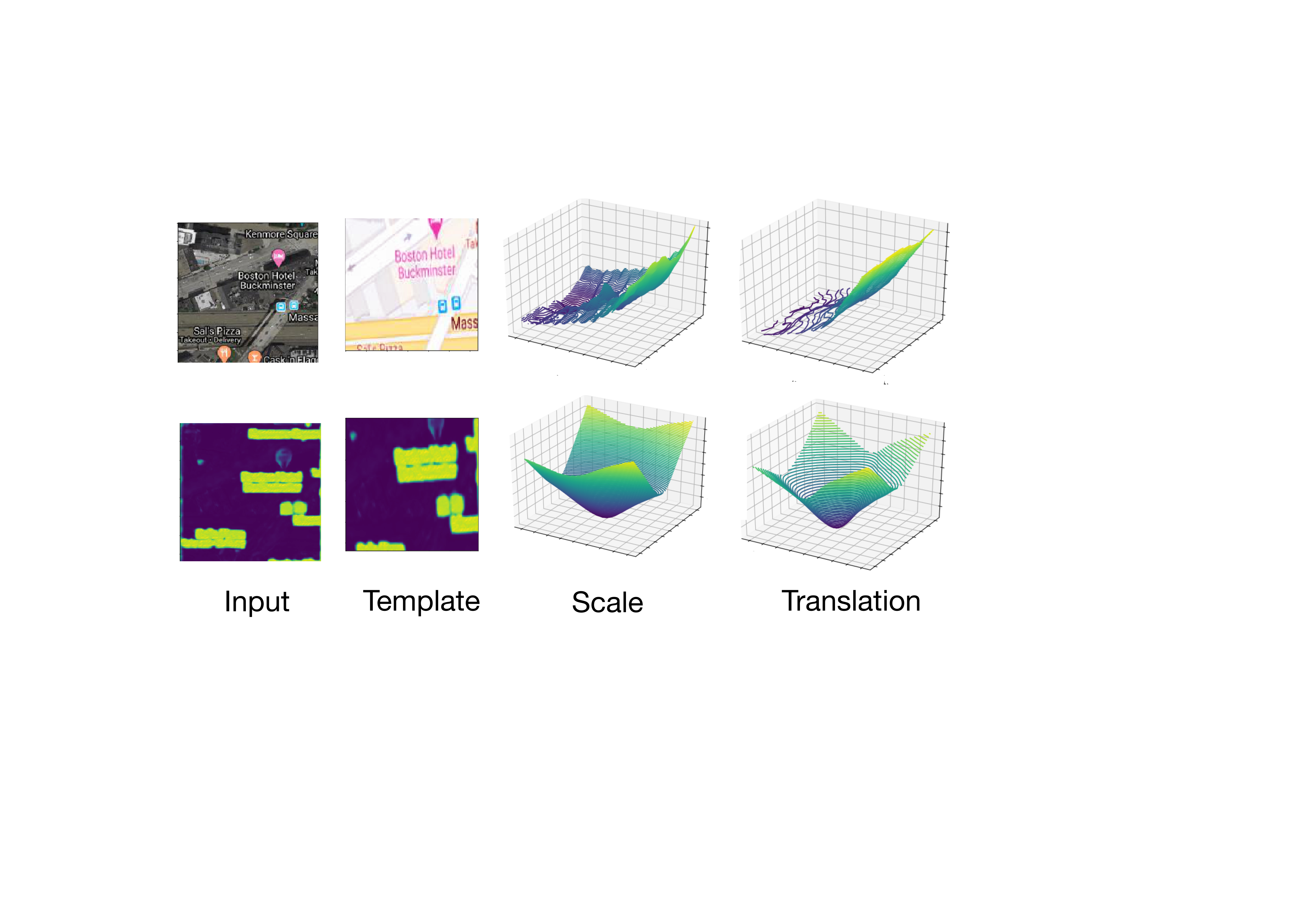}   
  \caption{{\bf (Top)} The LK objective function on the original multimodal image pair from Google Map and Satellite. {\bf (Bottom)} The objective function on our DLKFM. We respectively perturb the scale and translation in two directions around the ground truth whereas keeping the other six parameters fixed.}
\label{fig:8}
  \vspace{-1.2em}
\end{figure}

Our special designed loss function has two terms. The first term helps two feature maps keep the brightness consistency, which has been presented in Fig. \ref{fig:1}. The second term helps the convergence of Eq. \ref{eqn:braodcast_equation} by shaping the landscape of optimization function around the ground truth value $\bar{P}$, that is $\bar{P}=[\bar{P}_{11},\bar{P}_{12},\bar{P}_{13},\bar{P}_{21},\bar{P}_{22},\bar{P}_{23},\bar{P}_{31},\bar{P}_{32}]$. To verify it, we keep six parameters fixed, and perturb the other two around the ground truth. On the image coordinate, we perturb scale $(\bar{P}_{11},\bar{P}_{22})$ in two directions and translation $(\bar{P}_{13},\bar{P}_{23})$ in two directions, shown in Fig. \ref{fig:8}. The sample is from our multimodal Google Map and Satellite dataset. For original image pair, we can see the $\Delta P$ has no chance to converge as the $\bar{P}$ is even not the minimum. On the contrary, the landscape of the objective function on learned DLKFM shows a smooth surface with the minimum at $\bar{P}$. Therefore, by minimizing the objective, $\Delta P$ will keep moving toward the ground truth $\bar{P}$.




\section{Conclusion}
This paper proposes a generic method to align multimodal images by extending Lucas-Kanade on feature maps. Solving this fundamental challenge has the potential to provide solutions for many computer vision applications in the filed like UAV or remote sensing. Two specific techniques have been developed as part of our method. Since how to extract meaningful features and how to shape the objective function are both widely existing challenges, we believe those two technical contributions can give a hint to those related computer vision or machine learning tasks.

{\small
\bibliographystyle{ieee_fullname}
\bibliography{egbib}
}

\end{document}